\documentclass[11pt,a4paper]{article}

\usepackage{extract}
\usepackage{fancyhdr}
\usepackage[latin1]{inputenc}
\usepackage{amsmath}
\usepackage{amsfonts}
\usepackage{amssymb}
\usepackage{graphics}
\usepackage[hmargin=1.5cm,vmargin=2cm]{geometry}
\usepackage{multirow}
\usepackage{wrapfig}
\usepackage{epsfig}
\usepackage{framed}
\usepackage{textcomp}
\usepackage{color}
\usepackage[table]{xcolor}
\usepackage{booktabs}
\usepackage{pdflscape}
\usepackage[export]{adjustbox}
\usepackage{listings}
\usepackage[framemethod=tikz]{mdframed}
\usepackage[colorinlistoftodos]{todonotes}
\usepackage{booktabs}
\usepackage{siunitx}
\usepackage{hyperref}
\usepackage{xspace}
\usepackage{lscape}
\usepackage{adjustbox}
\usepackage{soul}
\usepackage{algorithm}
\usepackage[noend]{algpseudocode}
\usepackage[export]{adjustbox}
\usepackage{amsthm}
\usepackage{xfrac}
\usepackage[]{graphicx}
\usepackage[]{color}
\usepackage[all]{nowidow}
\definecolor{fgcolor}{rgb}{0.345, 0.345, 0.345}

\usepackage{todonotes}
\usepackage{multicol}
\usepackage{xcolor}
\usepackage{soul}
\usepackage{cite}
\usepackage{booktabs}
\usepackage{siunitx}
\usepackage{amsmath}
\usepackage{amssymb}
\IfFileExists{upquote.sty}{\usepackage{upquote}}{}

\begin{document}
\title{\bfseries ToxicBlend: Virtual Screening of Toxic Compounds \\
                 with Ensemble Predictors}
\author{Mikhail Zaslavskiy, Simon J\'{e}gou, Eric W. Tramel, \& Gilles Wainrib\\ 
        Owkin, Inc. \\
        New York City, New York}
\maketitle

\abstract{
  Timely assessment of compound toxicity is one of the biggest
  challenges facing the pharmaceutical industry today. A significant proportion of 
  compounds identified as potential leads are 
  ultimately discarded due to the toxicity they induce. 
  In this paper, we propose a novel machine learning
  approach for the prediction of molecular activity on  
  ToxCast targets.  We combine extreme gradient boosting with fully-connected
  and graph-convolutional neural network architectures trained on
  QSAR physical molecular property descriptors,
  PubChem molecular fingerprints, and SMILES sequences. 
  Our ensemble predictor leverages the strengths of each individual technique, 
  significantly outperforming existing
  state-of-the art models on the \textsc{ToxCast} and \textsc{Tox21} toxicity-prediction
  datasets. We provide free access to molecule toxicity prediction using our
  model at \url{http://toxicblend.owkin.com}.
}

\vspace{4ex}\noindent\textbf{Keywords ---} Toxicity prediction, Ensemble models, Virtual screening 

\section{Introduction}
\label{sec:introduction}

In silico models for molecular activity prediction have become critical for drug
discovery and other biotechnology industries~\cite{Tanrikulu:2013aa}.  These
methods prove very important for the prediction of intrinsic chemical properties
(such as solubility~\cite{Palmer:2007aa} or acidity constant~\cite{Rupp:2011aa})
or target affinity prediction~\cite{Ung:2016aa}, including potential toxicity
effects, a crucial step in the development of new drugs.

Over the years, two complementary computational approaches have been developed to 
achieve this goal. The first approach, physical modeling, focuses on computational 
simulations of physical molecular interactions. This approach requires the
knowledge about the nature of the target and its well-described 3D 
structure.
Subsequently, significant computational resources are required for running the
simulation, as the space of possible configurations is vast.  While this
approach provides accurate descriptions of molecular dynamics, these two
requirements prohibit large-scale deployment of this technique to ameliorate the
costs associated with physical high-throughput screening (HTS). 

In the second approach, one instead seeks to construct a predictive model
learned from a large set of experimental measurements via statistical modeling
and machine learning.  While the training of such models may be costly and
time-consuming, their usefulness in predicting the toxicity of novel compounds 
can be observed in their scalability, as they allow for virtual screening of vast
chemical databases, even when structural information is not available.

One of the major limiting factors in the successful deployment of advanced 
machine learning methods for toxicity prediction is the requirement for large
datasets from which to train such complex models. 
To foster further development of machine learning techniques, 
multiple public initiatives~\cite{Subramanian-A:2017aa} have been 
launched to compile datasets with wide coverage of compounds and their
properties.
One such project is the \textsc{ToxCast} project~\cite{Palmer:2007aa}
supported by the U.S. Environmental Protection Agency (EPA), which offers the
scientific community unprecedented datasets of observed in vitro interactions
between thousands of small molecules and known toxicity-related targets.  The
\textsc{ToxCast} project seeks to better understand the underlying mechanisms 
of action of various toxic chemicals, and to provide support for predictive 
models that could be used to prioritize potential toxicants for in vivo testing.   

As witnessed in the \textsc{Tox21} Challenge, a spin-off of the \textsc{ToxCast}
project organized in 2014~\cite{Huang:2017aa}, techniques based around deep neural
networks (DNNs), are able to provide state-of-the-art performance for virtual
toxicity screening~%
\cite{koutsoukas2017predictive,  huang2017tox21, ribay2016predictive,
drwal2017molecular, stefaniak2017prediction, huang2016tox21challenge}.
The winning solution of~\cite{mayr2017deeptox} utilized an
ensemble of various classifiers built upon many thousands of chemical 
descriptors.

Subsequently, a wide range of new descriptors and chemical fingerprints
have been used in conjunction with machine learning approaches for the 
prediction of various properties of molecules
\cite{Ramsundar:2015aa, Dahl:2014aa, unterthiner2014deep, wallach2015atomnet,lusci2013deep, ma2015deep}. 
More recently, several papers have explored methods such as graph convolutional networks 
(GCNs)
\cite{duvenaud2015convolutional, wu2017moleculenet, gomes2017atomic, kearnes2016molecular}, 
bypassing conventional chemical fingerprinting and working directly on the molecular graph
structure, leading to state-of-the-art performance across a wide range of 
molecular prediction tasks, as reported in~\cite{Wu:2017}. 
Such approaches draw inspiration from convolutional neural networks for images,
even directly from \emph{drawings} of molecules as in
\cite{goh2017chemception}.

In this work, we take a pragmatic approach, proposing a systematic ensembling 
method for virtual toxicity screening.
We combine various molecular descriptors with three different 
predictive modeling approaches: gradient tree boosting, DNNs, and GCNs. 
The aim of this work is to establish state-of-the-art
predictors for 617 \textsc{ToxCast} targets as well as twelve \textsc{Tox21} and
to provide the scientific community with a ready-to-use prediction model for toxicity
pre-screening through our web server ToxicBlend \url{http://toxicblend.owkin.com}.

To show the advantage of the proposed model, we report objective results against baseline
virtual toxicity screening methods such as those implemented in DeepChem~\cite{Wu:2017}
using cross-validation performed via three different splitting scenarios: by index,
by scaffold, as well as by random selection.

\section{Materials}
\subsection{Datasets}
\label{sec:datasets}
In our experiments, we use the \textsc{Tox21} and \textsc{ToxCast} datasets obtained 
from the DeepChem package\cite{Wu:2017}. The \textsc{Tox21} dataset was constructed for the
Toxicity prediction challenge~\cite{Huang:2017aa}, organized by the National Center 
for Advancing Translational Sciences (NCATS) at the National Institutes of Health (NIH). 
The \textsc{ToxCast} dataset, constructed for the Toxcicity Forecaster program, 
is much borader in scope, containing many hundreds of toxicity assay targets.

\paragraph{\textsc{Tox21.}}%
The \textsc{Tox21} dataset consists of toxicity assays conducted over 8,000 chemical compounds 
on twelve different targets. 
Seven assays represent the response of five nuclear receptors: 
the androgen and estrogen receptors, for both the general and ligand binding domain, as well as 
the aryl hydrocarbon, glitazone, and aromatase receptors. The five remaining assays measure the
following cell stress responses: oxidative stress (ARE), genetic stress (ATAD5), heat shock
response (HSE), mitochondrial function disruption (MMP), as well as DNA damage and 
other miscellaneous cellular stress (p53).

\paragraph{\textsc{ToxCast.}}%
The \textsc{ToxCast} dataset is similar in nature to \textsc{Tox21}, but with a much wider range 
of performed toxicity assays. 
Table~\ref{tab:toxcast_assays} provides the distribution of \textsc{ToxCast} assays per target 
family type. In total, there are 617 assays measuring the toxicity effects of more than 8,000 
chemical compounds.   

\begin{table}[ht]
\begin{center}
\begin{tabular}{lrlr}\toprule
  {Family}                  & {Count} & {Family}             & {Count} \\ \midrule
  {Nuclear Receptor}        & 115     & {Oxidoreductase}     & 12 \\
  {DNA Binding}             & 77      & {Kinase}             & 11 \\
  {Background Measurement}  & 62      & {Ion Channel}        & 9\\
  {GPCR}                    & 55      & {Phosphatase}        & 7\\
  {Cytokine}                & 55      & {Esterase}           & 6\\
  {Cell Cycle}              & 51      & {Transferase}        & 5\\
  {Cell Morphology}         & 30      & {Growth Factor}      & 4\\
  {CYP}                     & 26      & {Misc. Protein}      & 3\\
  {Cell Adhesion Molecules} & 21      & {Hydrolase}          & 3\\
  {Protease}                & 17      & {Protease Inhibitor} & 2\\
  {Steroid Hormone}         & 14      & & \\
  {Transporter}             & 12      & {\textbf{Not defined}}& 20\\
  \bottomrule
\end{tabular}
\end{center}
\caption{Number of toxicity assays performed per target family for the \textsc{ToxCast} dataset.
         \label{tab:toxcast_assays}} 
\end{table}

\subsection{Dataset Stratifications}
\label{sec:splits}
To estimate the performance of individual models, as well as the final 
blended predictor, we perform three types of cross-validation experiments based
on random, scaffold and index splits, as in~\cite{Wu:2017}.
For random splitting, all ligands were divided randomly into three groups: 
training (80\%), validation, (10\%) and test sets (10 \%). 
When using scaffold splitting, all ligands were grouped according to their
scaffolds, such that ligands with the same scaffold were always assigned to 
the same set (training, validation, or test). Index based splitting follows the natural order of rows in the
dataset, it uses first 80\% rows as the training set and the following 10\% and 10\% as
validation and test sets.

\section{Molecular Feature Extraction}
\label{sec:features}
The final predictive model is built on top of several individual machine
learning models trained on various feature subsets. We tested different 
molecular fingerprint families such as PubChem fingerprints, 
extended connectivity fingerprints~\cite{Rogers:2010aa}, 
DFS fingerprints~\cite{Ralaivola:2005aa},
molprint 2D fingerprints~\cite{Bender:2005aa}, 
topological autocorrelation keys~\cite{Schneider:1999aa}, 
Shannon entropy descriptors~\cite{Gregori-Puigjane:2006aa}, 
and several molprint 3D fingerprints implemented in 
java Compound Mapper~\cite{Hinselmann:2011aa}: geometrical autocorrelation keys, 
3D atom triplets~\cite{Mahe:2006aa} and 3D molprint-like fingerprints. 

Another, completely different,
source of information about molecules was derived using computational
models for the estimation of various molecular properties ranging from simple 
(e.g. total molecular mass, number of atoms, number of chemical bonds, etc.), 
to complex, such as molecular solubility and polarity, which are 
known to be
directly related to molecular activity potential~\cite{Alsenz:2007aa}. In addition to these
two groups of descriptors, we also tested the capacity of machine learning
models to work directly on the molecule SMILES, themselves. Although SMILES
representations can be seen to be overly complicated for ML techniques, compared to
molecular graphs for instance, 
we are able to show that models trained directly on SMILES features can be competitive
with those trained on molecular fingerprint representations for both 
\textsc{Tox21} and \textsc{ToxCast}. For the individual models we train, we 
utilize physical ligand descriptors, PubChem molecular fingerprints, and 
SMILES $n$-grams, which we describe in detail below.

\paragraph{Physical Ligand Descriptors (PLD).}%
The first set of 47 features correspond to various QSAR molecular descriptors computed using
the chemistry development kit~\cite{Steinbeck:2003aa}. These descriptors represent
physico-chemical properties of molecules such as mass, atom count, solubility,
and polarity, among others. We refer the reader to Sec.~\ref{sec_list_physical_descriptors} 
for a complete list of extracted features. 

\paragraph{PubChem Molecular Fingerprints (PCFP).}%
The PubChem fingerprint is a 881-bit descriptor indicating the presence or absence 
of a predefined list of 881 molecular subgraphs and subgraph families within a given
compound.\footnote{ftp://ftp.ncbi.nlm.nih.gov/pubchem/specifications/pubchem\_fingerprints.txt}. 

\paragraph{SMILES $n$-grams (SNG).}%
SMILES $n$-gram features consist of counts of SMILES sub-sequences, up to $n$ characters.
In our experiments, we used $n \in \{3, 4\}$. These features can be seen as a simplified version of
molecular-graph fingerprints, where fingerprints are defined by the canonical SMILES
representation~\cite{Steinbeck:2003aa}.

\section{Toxicity Prediction Models}
\label{sec:models}
We tested every molecular feature subset in conjunction with the a few machine
learning models trained in a multi-task setting, namely:
extreme gradient boosting trees (Xgboost)~\cite{Chen2016},
DNNs \cite{CBH2015}, as well as GCNs as implemented in the 
DeepChem package~\cite{Wu:2017}. All model hyper-parameters were optimized using
the validation set. In addition to testing the prediction performance of our
final blended model using these models, we also attempted to blend additional
baseline classifiers such as Lasso, ride regression, random forests \cite{Ho1995},
and K-nearest neighbors. However, we found empirically that these simple 
methods did not provide a significant improvement in toxicity prediction 
accuracy.

\subsection{Multi-task extreme gradient boosting}
Our multi-task gradient boosting model is trained on an 
entire dataset simultaneously by stacking all task-specific data 
together and adding a descriptor representing the identity of the
corresponding task. This representation allows the gradient boosting algorithm to
leverage information from different tasks and to learn characteristics indicating ligand
toxicity across multiple tasks. At the same time, the algorithm can build task
specific rules by combining the task identity descriptor with the corresponding ligand
features. 

To train a gradient boosting model, we used the Xgboost package~\cite{Chen2016}.
For each cross-validation iteration, we optimized three hyper-parameters 
using the held-out validation split: 
i) the number of rounds, up to 20k rounds with
an early stopping triggered by validation performance, 
ii) the depth, up to 20, 
and  iii) the shrinkage parameter, which was selected from the set
of values $\left\{ 1{\rm e}-2, 5{\rm e}-3, 1{\rm e}-3, 5{\rm e}-4 \right\}$.

\subsection{Multi-task neural networks}
Deep learning methods have proved to be very efficient for the toxicity prediction task
were essential in the winning methods of 
the \textsc{Tox21} challenge~\cite{Mayr:2016aa,Capuzzi:2016aa}. 
Multi-task neural networks are particularly effective for molecular
inputs, as shown in~\cite{Ramsundar:2015aa}. Instead of building $T$ individual
networks to tackle $T$ different tasks, we use a single network with $T$ outputs. 
This construction permits the use all available information on the input 
molecule within a single network, enriching intermediate representations.

As in ~\cite{Mayr:2016aa}, all neural networks architectures were manually 
optimized for each evaluated dataset and each set of molecular features.
In contrast with the single-layer multi-task NN demonstrated in~\cite{Duvenaud:2015aa},
we optimized architectures over a set of shallow NN models 
consisting of $L = \{2, 3\}$ layers, each of which may contain 
$M = \{256, 516, 1024\}$ neurons. 
All networks use ReLU activations\footnote{${\rm ReLU}(x) = \max(x,0)$}, 
with dropout regularization applied to each layer at a rate of $0.5$ to
prevent over-fitting.
In the case of NNs trained on PCFP inputs, we used a dropout rate of $0.1$
in order to account for the extreme sparsity of these features.
All networks were trained to minimize the average binary cross-entropy 
over the set of $T$ targets,
\begin{equation}
\mathcal{L}(\mathbf{y}, \widetilde{\mathbf{y}}; \Theta) \triangleq
\frac
{\sum_{t=1}^{T} \sum_{i=1}^{N} \delta(1 + y_{ti}) \times
\left[ y_{ti} \log \widetilde{y}_{ti} + (1-y_{ti})\log(1-\widetilde{y}_{ti})\right]}
{\sum_{t=1}^{T} \sum_{i=1}^{N} \delta(1 + y_{ti})},
\label{eq:loss}
\end{equation}
where $\delta(\cdot)$ is the Dirac delta function such that $\delta(0) = 1$ and $0$ everywhere 
else, and $\widetilde{\mathbf{y}}=[\widetilde{y}_{ti}; \forall t, i ]$ are the labels inferred
from the NN with parameters $\Theta$ for the given mini-batch training samples. 
The ground-truth labels $\mathbf{y} = [y_{ti}; \forall t, i ]$ are assigned $1$ if the ligand $i$ 
is active for the target $t$, $0$ if it is inactive, and $-1$ if the information is not available. 
The value $N$ is the mini-batch size, $N = 512$ for our experiments. 
The model parameters $\Theta$ were optimized to minimize 
\eqref{eq:loss} using Adam~\cite{KB2014}. 
We used early stopping after 25 epochs when no improvement was observed on the 
loss of the validation set. For our implementation, we use the
Keras~\cite{Cho2015} framework in conjunction with 
Tensorflow~\cite{tensorflow2015-whitepaper}.

\subsection{Graph Convolutional Models}
GCNs \cite{Duvenaud:2015aa}, 
can be seen as a more general version of molecular circular
fingerprint~\cite{Glen:2006ab}, where fingerprints are learned directly from 
the dataset. For more details on this approach as applied to molecular data, 
we refer the reader to \cite{Wu:2017, Duvenaud:2015aa}. In our experiments, we
train GCNs using the DeepChem package \cite{Wu:2017}.

\section{Model Blending}%
\label{sec:blend}

For state-of-the-art toxicity prediction, we propose a final predictor which
is built on top of different models trained on the three feature subgroups we detail
at the end of Sec.~\ref{sec:features}. During experimentation, we found that 
this approach is more efficient, and less prone to overfitting,
compared to a single model trained over all available extracted molecular features.

We built our blended predictor using  multi-task gradient boosting
with an additional constraint on the monotonous dependency between the individual
model predictions and the final combined score~\cite{Ridgeway:2007aa}. 
This constraint helps avoid overfitting and also promotes an efficient 
combination of prediction models, even when the number of training samples is small.

\section{Results}%
\label{sec:results}
For each of the individual models, validation splits are used for hyper-parameter tuning.
Toxicity predictions on the held-out test sets were used for the assessment of final
model performance.
Our proposed blended predictor was trained using individual model predictions on
the validation splits.               

\begin{figure}[t]
\centering 
\includegraphics[width=.3\linewidth]{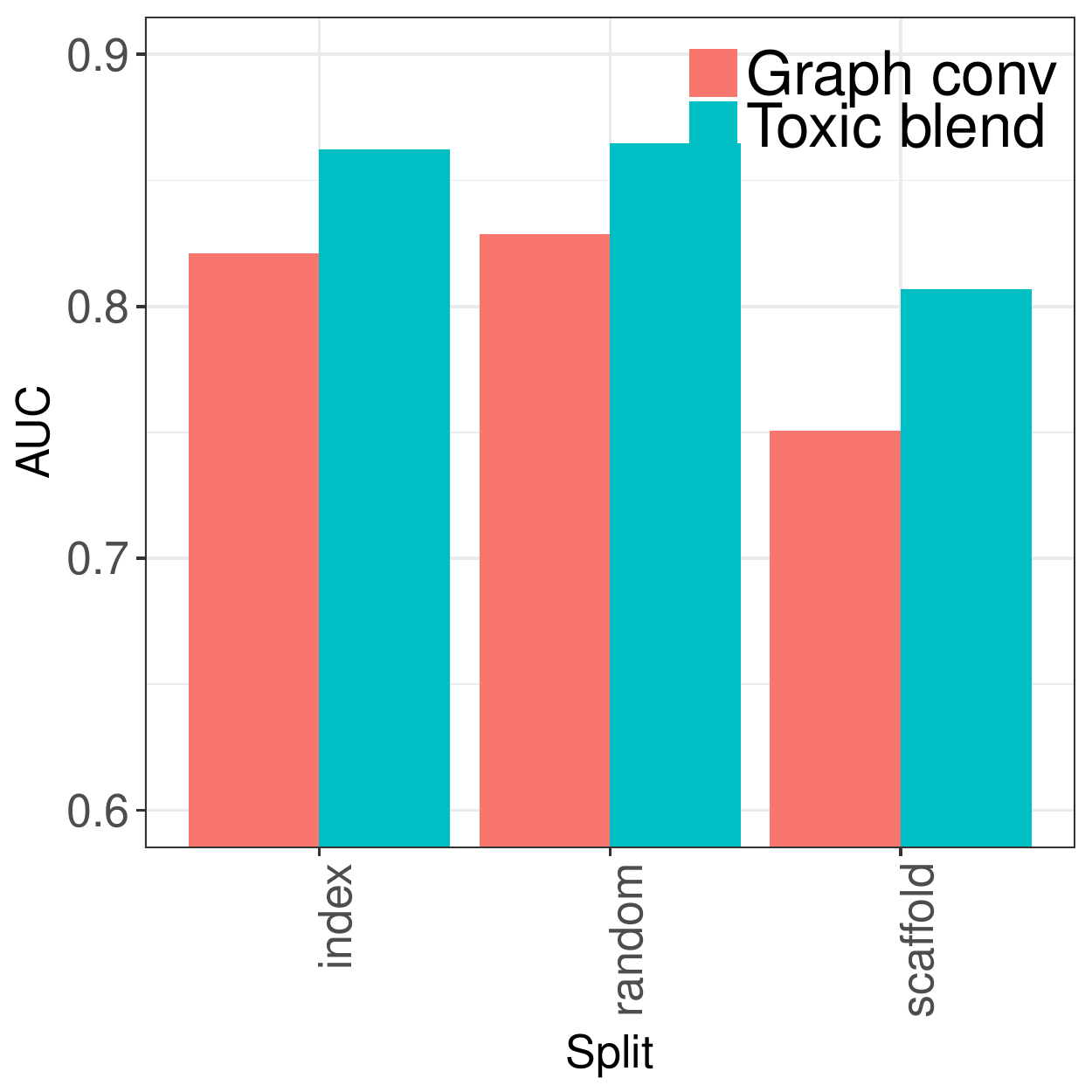} 
~
\includegraphics[width=.3\linewidth]{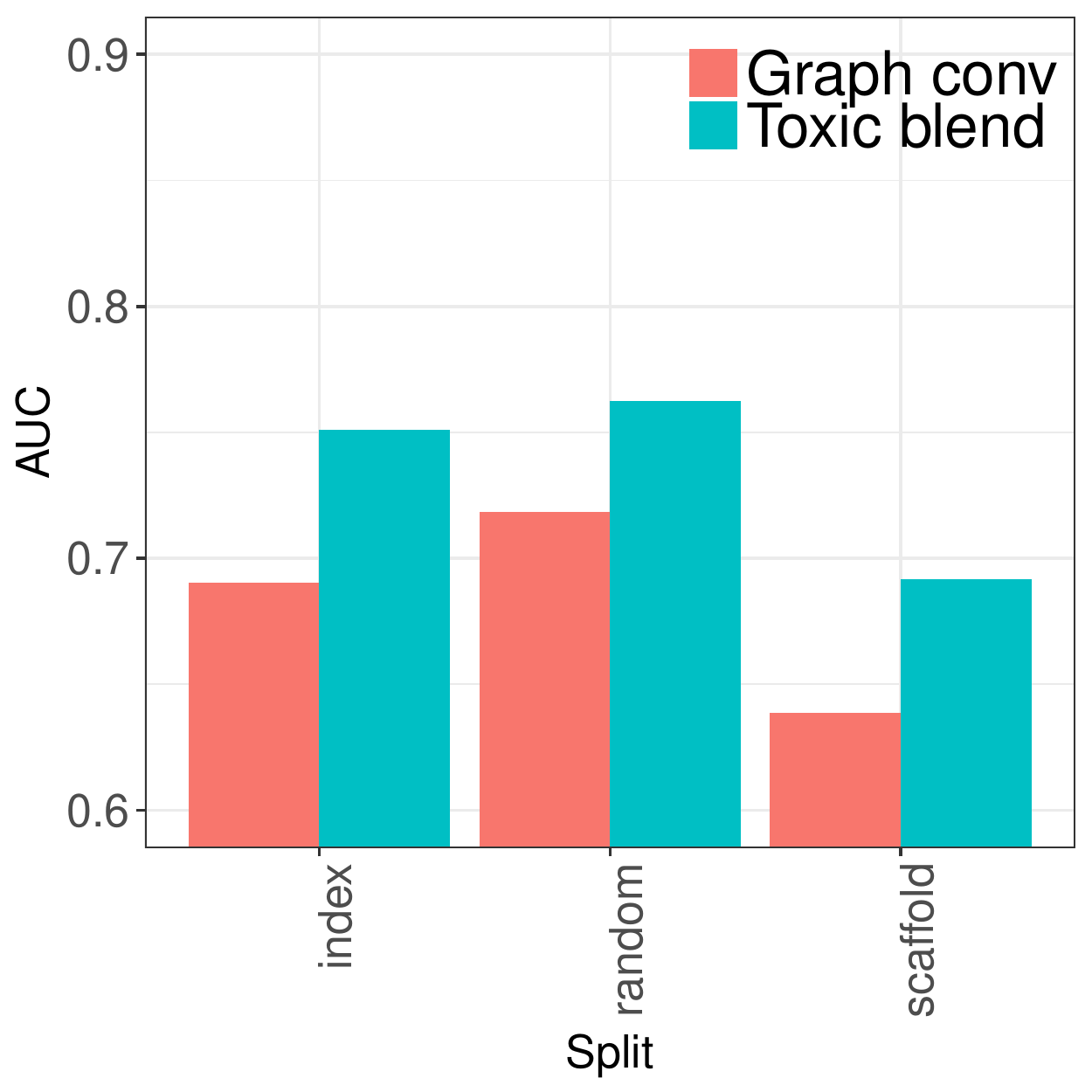} 
~
\includegraphics[width=.3\linewidth]{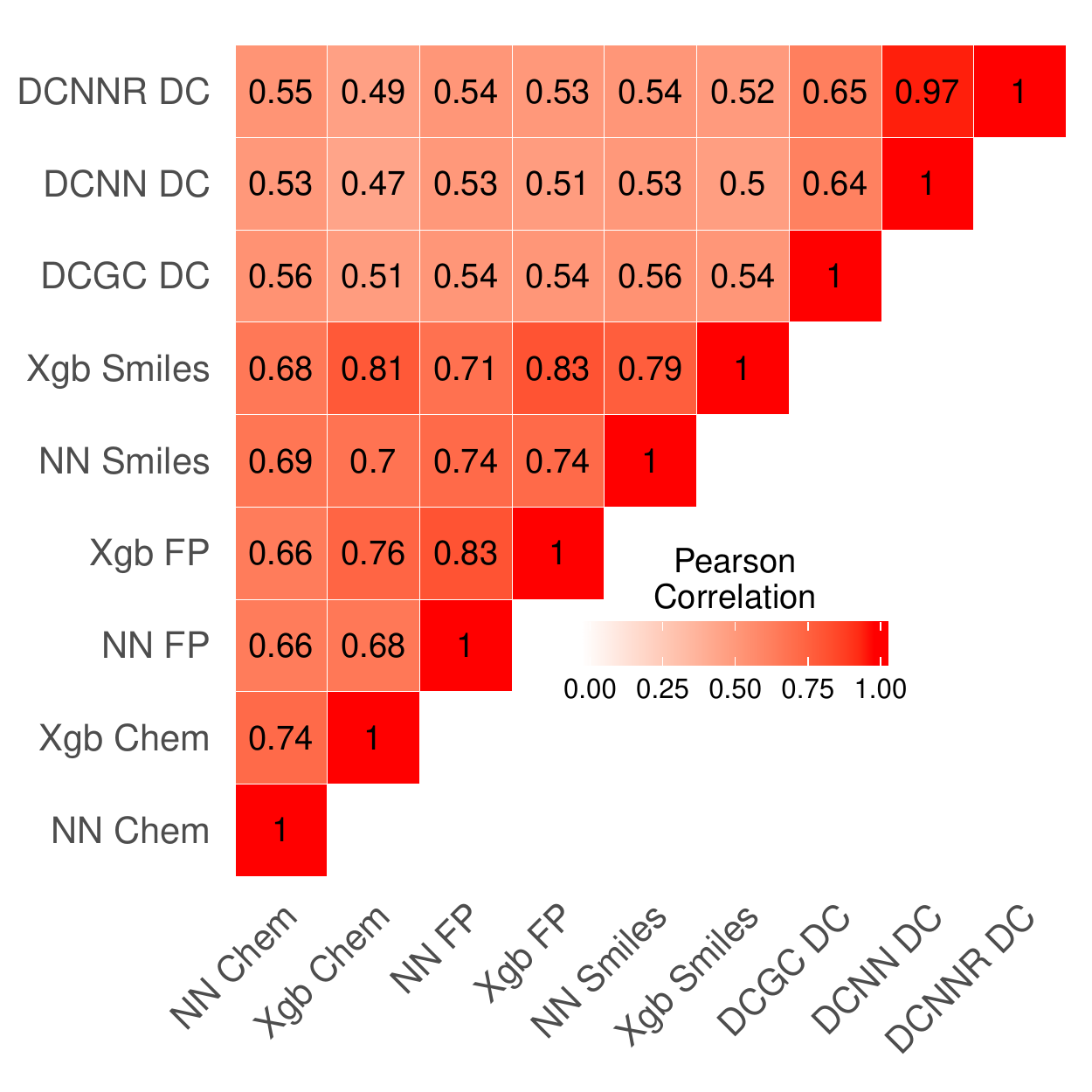} 
  \caption{%
           AUC comparison between state-of-the-art toxicity prediction with a
           single GCN, and the proposed blended ensemble of multiple 
           approaches, shown over different cross-validation split approaches,
           for the \textsc{Tox21} (\textbf{left}) and \textsc{ToxCast} (\textbf{middle})
           datasets. \textbf{Right:} Correlation coefficients between individual 
           prediction models in the blended ensemble, averaged over different 
           tasks in the \textsc{Tox21} dataset.}

\label{fig_all}
\end{figure}

Table~\ref{tab_results_all_blend} presents the performance of all individual models, as
well as our blended predictor. Here, the presented AUC values are averages over ten
independent trials for both the random and scaffold splits. For the index split, only
a single trial is used. 
Of the individual models, Xgboost trained on chemical descriptors provides the best 
AUC performance on random splits.
This result suggests that chemical properties estimated by computational models
can provide rich features for accurate toxicity prediction.
The performance gap grows when we split ligands by scaffold, indicating that the
chemical descriptors better describe cross-scaffold properties important for toxicity
prediction. Interestingly, models trained only on SMILES subsequences already provide 
enough information for machine learning models to generate competitive predictions. 
Finally, when we combine all methods together using our blended predictor, 
we observe a performance improvement of  0.5 to 2 AUC points.  Fig.~\ref{fig_all} 
compares this gain against the best-performing GCN model 
from the DeepChem package \cite{Wu:2017} for both datasets across the 
tested splitting strategies.

In Fig.~\ref{fig_all}, we also show the Pearson correlation coefficients 
calculated between individual models. 
Low correlation coefficients between the predictions of individual models 
indicates a diverse pool of predictors from which to build a blended model. Such
diversity generally leads to a robust final blend, as the failure of a 
particular indicator does not correlate with the failure of other indicators.
However, this diversity requires a thoughtful construction of the blending 
method. For example, the high variance between the performance of individual models
prevents the use of simple ensembling techniques such as prediction averaging.
Concretely, attempting such an averaging for random splits of the \textsc{Tox21}
dataset leads to an average AUC of 0.851: a level of performance worse than the
best performing individual model, as shown in Table \ref{tab_results_all_blend}.
In the case of random splits on \textsc{ToxCast},
prediction averaging provides an AUC of 0.754, a result better than the individual
models, but still less than our proposed blending approach (0.763).    

\begin{table}[ht]
\begin{center}
\small
\begin{tabular}{lccccccc}
  \toprule
  {Model} & Features &  \multicolumn{3}{c}{{\textsc{Tox21}}}  &  \multicolumn{3}{c}{{\textsc{ToxCast}}}  \\ 
  \cmidrule(lr){3-5}
  \cmidrule(lr){6-8}
  ~ & ~ & {\emph{index}} & {\emph{random}} & {\emph{scaffold}} & {\emph{index}} & {\emph{random}} & {\emph{scaffold}} \\
  \midrule
    {NN } & {PLD} & 0.821 & 0.821 & 0.751 & 0.732 & 0.716 & 0.657 \\     
    {NN } & {PCFP} & 0.831 & 0.837 & 0.761 & 0.730 & 0.738 & 0.670 \\ 
    {NN } & {SNG} & 0.821 & 0.839 & 0.770 & 0.722 & 0.739 & 0.663 \\ 
    \midrule
    {Xgb} & {PLD} & 0.842 & 0.853 & 0.803 & 0.742 & 0.740 & 0.680 \\ 
    {Xgb} & {PCFP} & 0.846 & 0.852 & 0.773 & 0.731 & 0.738 & 0.660 \\     
    {Xgb} & {SNG} & 0.854 & 0.852 & 0.777 & 0.738 & 0.742 & 0.668 \\ 
  \midrule
  {DCGC } & {DC} & 0.821 & 0.829 & 0.751 & 0.690 & 0.718 & 0.639 \\ 
  {DCNN } & {DC} & 0.785 & 0.787 & 0.724 & 0.651 & 0.676 & 0.614 \\ 
  {DCNNR} & {DC} & 0.784 & 0.790 & 0.730 & 0.654 & 0.683 & 0.614 \\ 
  \midrule
  %  {ToxicBlend} &  & \bf{0.862} & \bf{0.865} & \bf{0.807} & \bf{0.751} & \bf{0.762} & \bf{0.692} \\
  {Pred. Avg.} &  & 0.849 & 0.851 & 0.793 & 0.751 & 0.754 & 0.688\\
  {ToxicBlend} &  & \bf{0.866} & \bf{0.862} & \bf{0.807} & \bf{0.753} & \bf{0.763} & \bf{0.693} \\
  \bottomrule
\end{tabular}
\caption{AUC cross-validation performance scores, averaged over 10 independent
  random splits for random and scaffold splits. There is only one unique index
  split, the corresponding value is showed in the table.} 
\label{tab_results_all_blend}
\end{center}
\end{table}

\subsection{Per-target performance}
\begin{figure}
{\centering \includegraphics[width=.45\linewidth]{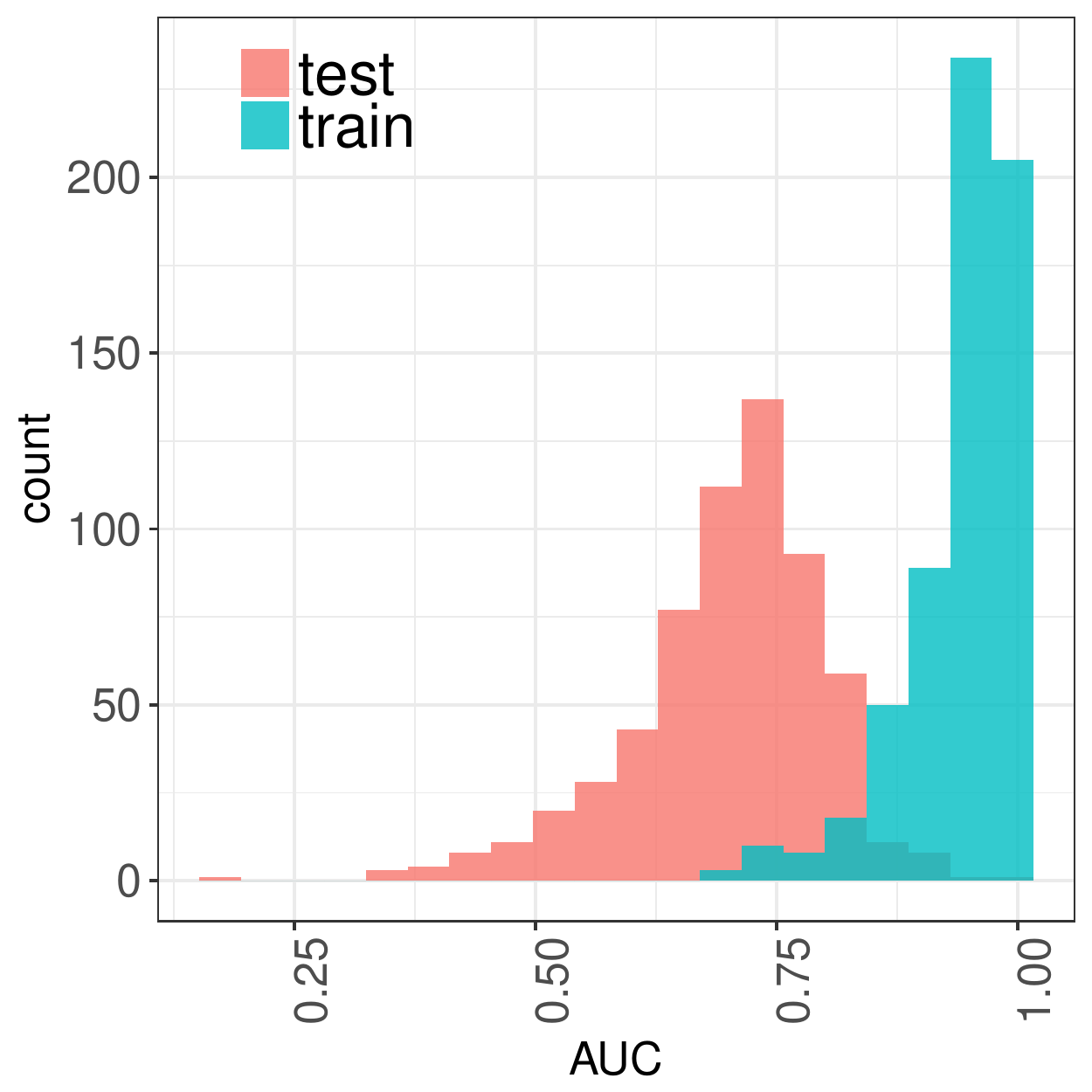} 
 \includegraphics[width=.45\linewidth]{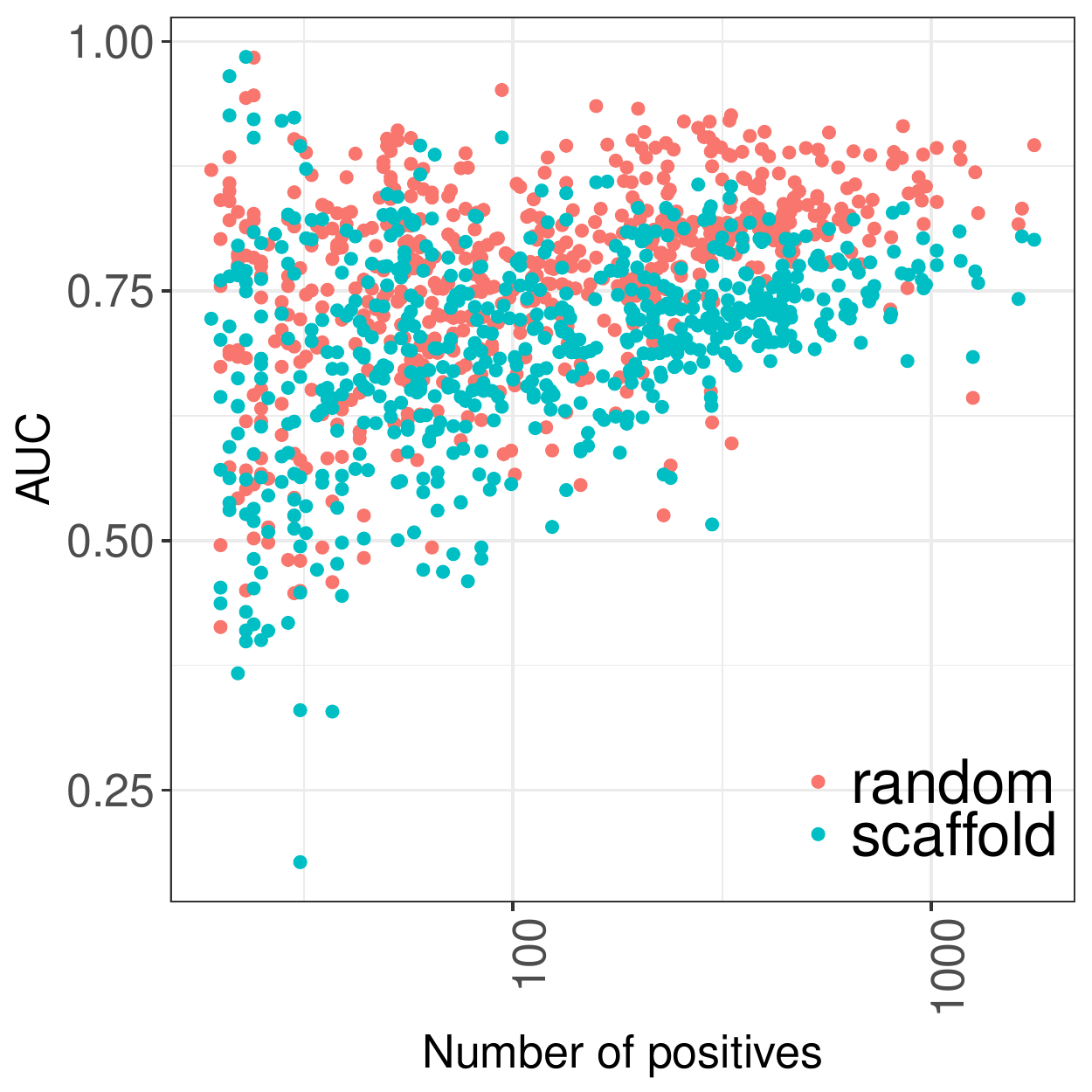} 
}
  \caption{%
  (Left): Distribution of train and test per target AUC scores, scaffold split on \textsc{ToxCast}
  dataset.  (Right):  Number of positive examples versus the target test AUC
  score in the case of random and scaffold splits.}  
\label{fig_per_target_performance_hist}
\end{figure}

The final scores in Tab.~\ref{tab_results_all_blend} are aggregated scores across
many different targets, some of which are harder to predict than others.
In Fig.~\ref{fig_per_target_performance_hist} (Left), we show 
the histogram of train and test AUC scores (scaffold split) on \textsc{ToxCast} 
dataset. 
While the training scores never fall below 0.7, 
on the test splits there are targets with an average AUC below 0.5,
and targets with an AUC above 0.9. 
We provide a list of the ten hardest and easiest to predict targets for 
\textsc{ToxCast} in Suppl. Table~\ref{tab_pertarget_auc}. 
There are several possible physical explanations for such a heterogeneity of 
performance scores across targets:
the capacity of the binding pocket to accept different types
of ligands, the level of noise in target specific assays, etc. 
We identified one important factor in determining per-target performance 
which arises simply from the data-collection process: the number of 
positive examples for each target. 
In Fig.~\ref{fig_per_target_performance_hist} (Right), we show the clear correlation between the
number of positive examples for each target plotted against and 
the per-target AUC achieved by our blended model. 
Targets with more than 350 positive examples, for instance, never fall below
an AUC of 0.64 or 0.69, for scaffold and random splits, respectively.

\subsection{Ligand feature importance}
\begin{figure}
  \centering 
  \includegraphics[width=.45\linewidth]{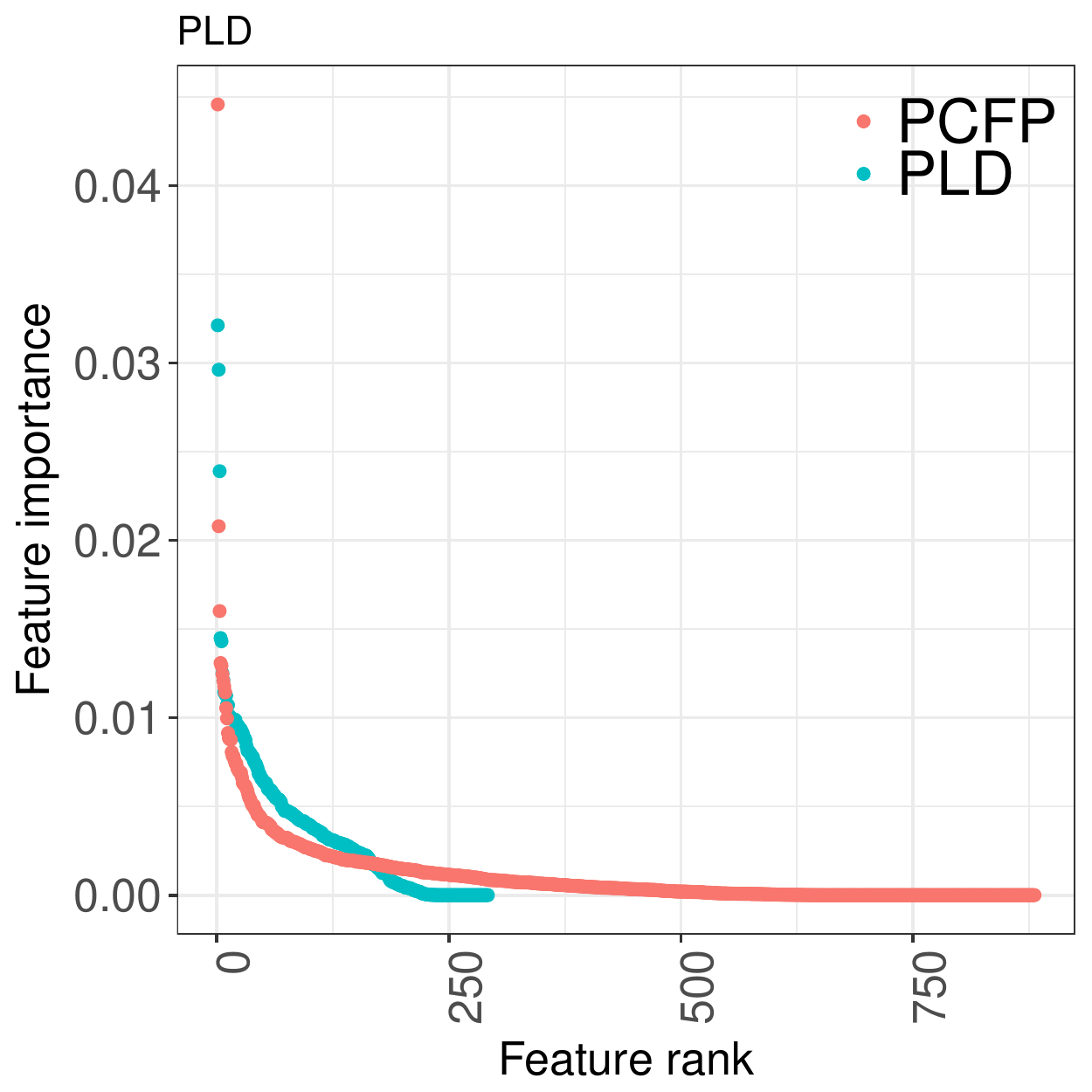} 
  \includegraphics[width=.45\linewidth]{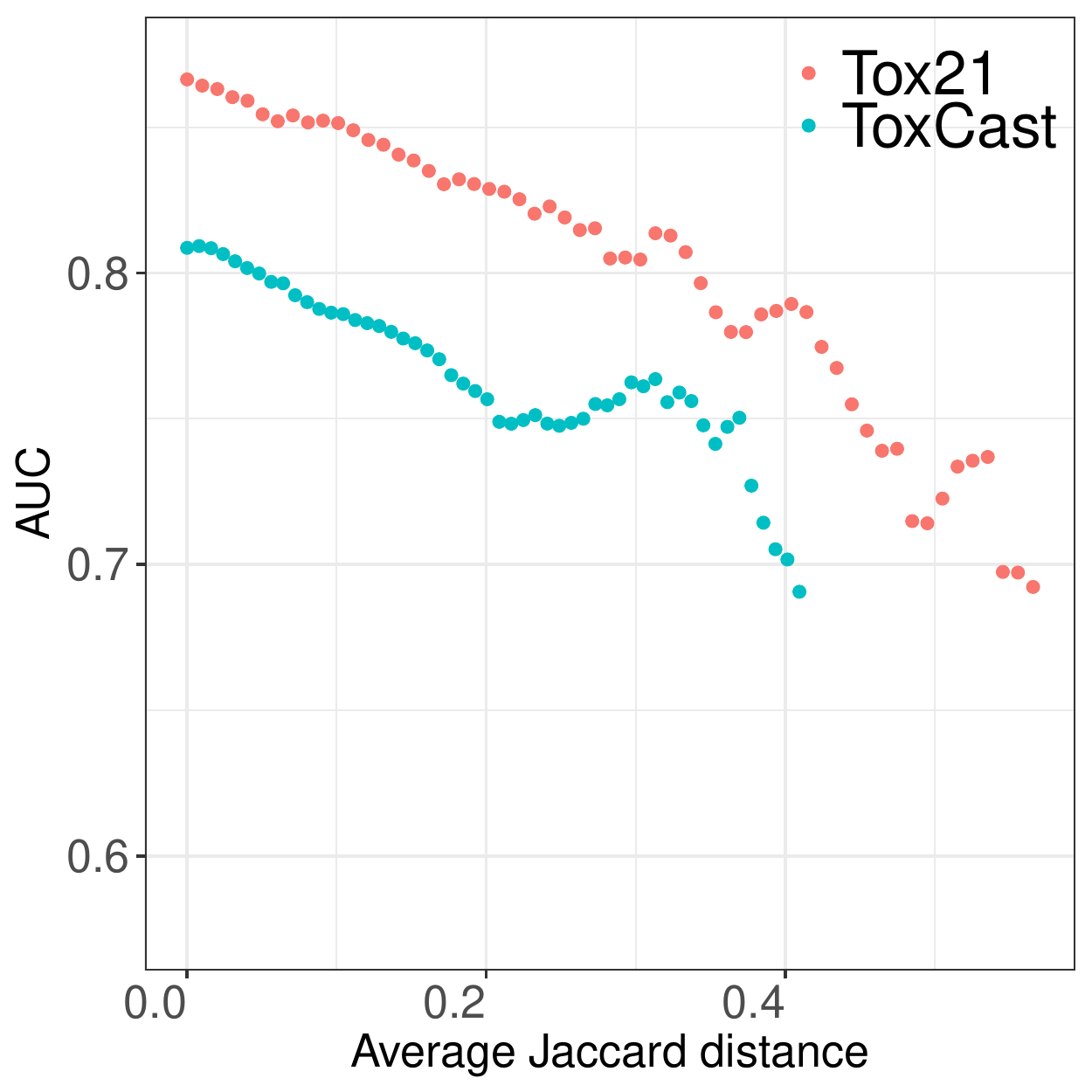} 
  \caption{%
  Left: Xgboost gain on \textsc{ToxCast} dataset for both PLD (\emph{left}) 
  and PCFP (\emph{right}) features. Right: Average AUC score against the average Jaccard distance to the
  training set.}  
  \label{fig_importance_cumulativegain}
\end{figure}

In this section we discuss  features that were identified as important by DL and
Xgboost models. Fig.~\ref{fig_importance_cumulativegain} (Left) shows the relative
ligand feature importance defined as the gain computed by gradient boosting
trees algorithm. 

For neural networks, we used a permutation feature importance technique. For each feature, we 
randomly shuffled the values associated to this feature and the $k$ most correlated features, 
and computed the average AUC degradation over 5 folds, for $k$ = 0, 5 and 10. The lower the AUC, 
the more important the feature.

Table~\ref{tab_toxcast_top10_features} lists the top ten features according to 
their importance ranking for both chemical and PubChem fingerprint descriptors. 
Interestingly, many of the most important fingerprint descriptors are
related to the size of the molecule, indicating that size characteristics might
be a critical indicator for toxicity prediction.
When trained on chemical descriptors, 
the individual Xgboost model prioritizes Broto-Moreau autocorrelation
features (ATSpx) built on atomic polarizabilities, Ghose-Crippen molar
refractivity (AMR), number of bonds, eigenvalue descriptors based on Burden
matrix, Wiener polarity number, and the solvent accessible surface areas of 
atoms (THSA and TPSA).    

\begin{table}[ht]\small
    \centering
        \begin{tabular}{rp{1.5cm}p{1.5cm}cp{5.1cm}p{5.1cm}}
      \toprule
       ~& \multicolumn{2}{c}{Chem}  & ~ & \multicolumn{2}{c}{PCFP}  \\ 
        \cmidrule(lr){2-3}
        \cmidrule(lr){5-6}
        ~&\multicolumn{1}{l}{\it Xgb} & \multicolumn{1}{l}{\it NN} & ~ & \multicolumn{1}{l}{\it Xgb} & \multicolumn{1}{l}{\it NN} \\ 
      \midrule
      1.& ATSp3 &    HybRatio & ~ &    {$\geq 1$ N} &  $\geq 16$ C \\ 
      2.& ATSp4 &    C1SP2 &  ~ &      O-C-C-C-C-C-C-C &  $\geq 1$ Sn \\ 
      3.& AMR &      Wnu1.unity & ~ &  $\geq 32$ C &   $\geq 2$ any ring size 6 \\ 
      4.& nB &       ATSm4 & ~ &       $\geq 16$ O &   C-C-C-C-C-C-C-C \\ 
      5.& BCUTp.1l & VP.3  & ~ &       $\geq 2$ unsaturated non-aromatic heteroatom-containing ring size 6 &   
                                        $\geq 2$ unsaturated non-aromatic carbon-only ring size 6 \\ 
      6.& ATSm1 &    MDEC.34 &  ~ &    $\geq 8$ O &  $\geq 3$ any ring size 6 \\ 
      7.& WPOL &     Weta1.unity & ~ & O=C-C-O &   $\geq 1$ Hg \\ 
      8.& THSA &     ATSp4 &~ &        C(\~{}N)(:C) &  Sc1c(Cl)cccc1 \\ 
      9.& TPSA &     ATSp1 &~ &        $\geq 4$ Cl &  SC1C(Cl)CCCC1 \\ 
      10.& ATSm2 &   ATSp3 & ~ &       $\geq 8$ Cl &  S(-O)(=O) \\ 
    \bottomrule
    \end{tabular}
    \caption{Ten most important chemical properties and PubChem fingerprint features according 
    to individual Xgboost and neural network classifiers trained on the 
    \textsc{ToxCast} dataset.} 
    \label{tab_toxcast_top10_features}
    \end{table}

\subsection{Reliability of the model}
Why should we trust our model? The cross-validated AUC estimate gives us a global confidence level 
per target, and for a given molecule, the probability estimated by the model measures the 
uncertainty of the prediction. However, both of these measures were obtained using the training 
distribution, which may not be representative of the distribution of molecules at test time.

To measure the distance from a new molecule $\mathbf{x}$ to the training distribution $X_{train}$, 
we use the average Jaccard distance on the $k=5$ nearest neighbors in the fingerprint space, 
$$ d(\mathbf{x},X_{train}) = \frac{1}{k}\sum_{\mathbf{x'} \in N_{train}(\mathbf{x}, k)}{1 - \frac{|\mathbf{x} \cap \mathbf{x'}|}{|\mathbf{x} \cup \mathbf{x'}|}}.$$

For each target, we compute the AUC score of the test molecules above a given distance threshold.
Averaged over 10 test folds (random splits), and all possible targets, we observed a that a higher 
distance threshold leads to a lower AUC score for both the \textsc{Tox21} and the \textsc{ToxCast} 
datasets, as shown in Figure~\ref{fig_importance_cumulativegain} (Right). This indicates that the model is less accurate on 
test molecules far from the distribution of training molecules. 

We also observe the same trend between the AUC score and the complexity of the test molecule, 
expressed as the count of ones in the fingerprint vector. We hypothesize that the higher diversity 
of complex molecules requires the model to be trained on more data to be accurate.

\subsection{Experimental variability}
Many biological experiments are subject to considerable variability, and it is
often impossible to build a perfect prediction for the simple reason that
the experiments themselves are not 100\% reproducible. 
Thus, it is useful to estimate the best-case toxicity prediction performance.  
The raw \textsc{Tox21} data, as published on 
NIH website\footnote{https://tripod.nih.gov/tox21},   
contains some instances of multiple measurements for the same molecule/target
couples. 
From these measurements, we observe significant heterogeneity between 
targets in terms of the result reproducibility. 
For example, 11\% of repeated measurements corresponding to NR-ER
contain contradictory results, while NR-PPAR-gamma have only 1\% of such
contradictory measurements. 
If we suppose that homogeneous repeated measurements are likely
to stay as they are, and contradictory measurements correspond to a random flip,
then an average (across all \textsc{Tox21} targets) AUC score that we could get by
predicting results of an experiment with an additional experiment would be 0.942.   

\section{Discussion}
In this paper, we present a new machine learning method for the prediction of
molecule activities in \textsc{Tox21} and \textsc{ToxCast} assays, 
namely, an approach to combine  multiple classifiers based on different molecular
descriptors. The new approach is implemented as a web server and can be accessed
at \url{http://toxicblend.owkin.com} and can be used to predict the toxicity of
chemical compounds from their SMILES.      

One of the main conclusions of our work is that it is beneficial to use known
molecular descriptors that were identified in the past as potentially important
for various biochemical properties of molecules, and it is important to combine
information sources of different kinds to improve the quality of predictions. An 
important direction for our future work is the incorporation of target-specific
information, such as target structure, in the prediction model. 
In experimentation (data not shown), 
we also used physical docking energy scores
computed by Smina~\cite{Koes:2013aa} to attempt to 
boost the final prediction performance on \textsc{Tox21} where 
seven out of twelve targets have known structures. However, it did not lead
to any significant improvements. One possible alternative strategy to simple
energy scoring is to use 
more elaborate structure-based descriptors representing
target pockets, and other target characteristics, and let ML algorithms determine
connections between ligand and target features. Such an approach would demand
high target diversity, thus an important direction for future work would
be to build an extensive representation for \textsc{ToxCast} targets
in order to evaluate the potential gain from such target descriptors. 
Mining scientific literature is another potentially interesting source of information.
This kind of approach can be seen as an automatic way to construct efficient molecular descriptors. 

To remain comparable with existing methods developed for toxicity prediction, we worked
only with the \textsc{Tox21} and \textsc{ToxCast} datasets, 
but an important future step would be
to use external data sources 
such as \textit{ChEMBL} for general ligand activity, \textit{BindingDB} for
structural information, or other existing datasets on molecule toxicity such as
\textit{RTECS}. 

\clearpage
\appendix
\section{Appendix}
\setcounter{table}{0}
\renewcommand{\thetable}{A.\arabic{table}}

\begin{table}[ht]\small
\centering
\begin{multicols}{2}
\begin{enumerate}
\item   ALOGPDescriptor
\item   APolDescriptor
\item   AminoAcidCountDescriptor
\item   AromaticAtomsCountDescriptor
\item   AromaticBondsCountDescriptor
\item   AtomCountDescriptor
\item   AutocorrelationDescriptorCharge
\item   AutocorrelationDescriptorMass
\item   AutocorrelationDescriptorPolarizability
\item   BCUTDescriptor
\item   BPolDescriptor
\item   BondCountDescriptor
\item   CPSADescriptor
\item   CarbonTypesDescriptor
\item   ChiChainDescriptor
\item   ChiClusterDescriptor
\item   ChiPathClusterDescriptor
\item   ChiPathDescriptor
\item   EccentricConnectivityIndexDescriptor
\item   FMFDescriptor
\item   FragmentComplexityDescriptor
\item   GravitationalIndexDescriptor
\item   HBondAcceptorCountDescriptor
\item   HBondDonorCountDescriptor
\item   HybridizationRatioDescriptor
\item   KappaShapeIndicesDescriptor
\item   KierHallSmartsDescriptor
\item   LargestChainDescriptor
\item   LargestPiSystemDescriptor
\item   LengthOverBreadthDescriptor
\item   LongestAliphaticChainDescriptor
\item   MDEDescriptor
\item   MannholdLogPDescriptor
\item   MomentOfInertiaDescriptor
\item   PetitjeanNumberDescriptor
\item   PetitjeanShapeIndexDescriptor
\item   RotatableBondsCountDescriptor
\item   RuleOfFiveDescriptor
\item   TPSADescriptor
\item   VABCDescriptor
\item   VAdjMaDescriptor
\item   WHIMDescriptor
\item   WeightDescriptor
\item   WeightedPathDescriptor
\item   WienerNumbersDescriptor
\item   XLogPDescriptor
\item   ZagrebIndexDescriptor
\end{enumerate}
\end{multicols}
\caption{List of physical descriptors}
\label{sec_list_physical_descriptors}
\end{table}
\clearpage

\begin{table}[ht]\small
\centering
\begin{tabular}{lcclcc}
  \toprule
  \multicolumn{3}{c}{Easiest Targets} & \multicolumn{3}{c}{Hardest Targets} \\
  \cmidrule(lr){1-3}
  \cmidrule(lr){4-6}
{\it Target} & {\it Test} & {\it Train} & {\it Target} & {\it Test} & {\it Train} \\ 
  \midrule
  TOX21\_AutoFluor\_HEPG2\_Media\_green & 0.98 & 1.00 & NCCT\_TPO\_GUA\_dn & 0.18 & 0.88 \\ 
  NVS\_NR\_rMR & 0.97 & 0.93                          & ATG\_Oct\_MLP\_CIS\_dn & 0.33 & 0.87 \\ 
  NVS\_GPCR\_hAT1 & 0.93 & 0.99                       & NVS\_ENZ\_hPTEN & 0.33 & 0.88 \\ 
  NVS\_NR\_hRAR\_Antagonist & 0.92 & 0.97             & NVS\_ENZ\_hMMP7 & 0.37 & 0.81 \\ 
  TOX21\_AutoFluor\_HEPG2\_Cell\_green & 0.92 & 1.00  & NVS\_ENZ\_hAurA & 0.40 & 0.95 \\ 
  NVS\_GPCR\_rNK3 & 0.92 & 0.93                       & BSK\_KF3CT\_MCP1\_up & 0.40 & 0.75 \\ 
  TOX21\_ESRE\_BLA\_ch1 & 0.90 & 1.00                 & NVS\_LGIC\_hNNR\_NBungSens & 0.41 & 0.80 \\ 
  TOX21\_PPARg\_BLA\_Agonist\_ch1 & 0.90 & 1.00       & ATG\_PXR\_TRANS\_dn & 0.41 & 0.94 \\ 
  APR\_HepG2\_NuclearSize\_24h\_dn & 0.90 & 0.99      & BSK\_KF3CT\_IP10\_up & 0.42 & 0.71 \\ 
  NVS\_ADME\_rCYP2C12 & 0.90 & 0.97                   & ATG\_TA\_CIS\_dn & 0.42 & 0.81 \\ 
  \bottomrule
\end{tabular}
\caption{
Top ten easiest and hardest to predict targets in \textsc{ToxCast} dataset.} 
\label{tab_pertarget_auc}
\end{table}

\bibliographystyle{ieeetr}
\bibliography{references}
\end{document}